\documentclass[runningheads]{llncs}
\usepackage[T1]{fontenc}
\usepackage{graphicx,verbatim}
\usepackage{lipsum}
\usepackage{graphicx,verbatim}
\usepackage{multirow}
\usepackage[table,xcdraw]{xcolor}
\usepackage[normalem]{ulem}
\useunder{\uline}{\ul}{}
\usepackage{booktabs}
\usepackage[colorlinks=true, linkcolor=blue, citecolor=black, urlcolor=magenta]{hyperref}
\usepackage{xspace}
\usepackage{amsfonts}
\newcommand{\ours}{\texttt{CoNNS}\xspace}
\usepackage{amsmath}
\usepackage{amssymb}
\usepackage{graphicx}
\usepackage{array}
\newcolumntype{C}[1]{>{\centering\arraybackslash}p{#1}}
\newcommand{\sbullet}{\,\vcenter{\hbox{\scalebox{0.6}{$\bullet$}}}\,}
\definecolor{positivecolor}{RGB}{199, 225, 181}
\definecolor{negativecolor}{RGB}{244, 182, 189}
\definecolor{ignoredcolor}{RGB}{231, 230, 230}
\definecolor{pendingcolor}{RGB}{255, 255, 255}
\usepackage{marvosym}

\usepackage{hyperref}
\usepackage{color}

\begin{document}
\title{Concept-Guided Noisy Negative Suppression \\for Zero-Shot Classification and Grounding \\of Chest X-Ray Findings}
\titlerunning{Concept-Guided Noisy Negative Suppression}
%

\author{
Chenyu Lian\inst{1,2} \and
Hong-Yu Zhou\inst{3} \textsuperscript{(\Letter)} \and
Chun-Ka Wong\inst{4} \and
Jing Qin\inst{1,2}\textsuperscript{(\Letter)}
}

%
\authorrunning{C. Lian et al.}
%
\institute{The Center for Smart Health, School of
Nursing, the Hong Kong \\ Polytechnic University, Hong Kong, China \\
\email{chenyu.lian@connect.polyu.hk, harry.qin@polyu.edu.hk} \and
Research Institute for Smart Ageing, the Hong Kong Polytechnic University, \\ Hong Kong, China \and
School of Biomedical Engineering, Tsinghua Medicine, Tsinghua University, \\ Beijing, China \\
\email{hongyu.zhou.ai@gmail.com} \and
Queen Mary Hospital, LKS Faculty of Medicine, The University of Hong Kong, Hong Kong, China. \\
\email{wongeck@hku.hk}
}
  
\maketitle              
\begin{abstract}
Vision-language alignment using chest X-rays and radiology reports has emerged as an advanced paradigm for zero-shot classification and grounding of chest X-ray findings.
However, standard contrastive learning typically treats radiographs and reports from different patients simply as negative pairs.
This assumption introduces \textbf{\textit{noisy negatives}}, as different patients frequently exhibit similar findings.
Such noisy negatives cause semantic ambiguity and degrade performance in zero-shot understanding tasks.
To address this challenge, we propose \ours, a \uline{co}ncept-guided \uline{n}oisy-\uline{n}egative \uline{s}uppression framework.
To support the negative suppression mechanism, unlike previous methods that use raw reports or templatized texts, we construct a hierarchical concept ontology using large language models.
The ontology structures 41 key clinical concepts by explicitly modeling presence, attributes (location and characteristics), and texts (evidential segment and presence statement).
Leveraging this ontology, we implement a cross-patient pair relabeling strategy comprising three steps: (1) \textit{Fine-Grained Breakdown} to categorize pairs based on finding presence; (2) \textit{Noisy Negative Filtering} to resolve semantic conflicts by removing false negatives; and (3) \textit{Hard Negative Mining} to identify subtle attribute discrepancies using a lightweight language model.
Finally, we propose a Concept-Aware NCE loss to align visual features with text while suppressing the identified noisy negatives.
Extensive experiments across multi-granularity zero-shot grounding tasks and five zero-shot classification datasets validate that \ours outperforms existing state-of-the-art models.
The code is available at \url{https://github.com/DopamineLcy/conns}.


\end{abstract}
\section{Introduction}
\begin{figure}[!t]
    \centering
    \includegraphics[width=\linewidth]{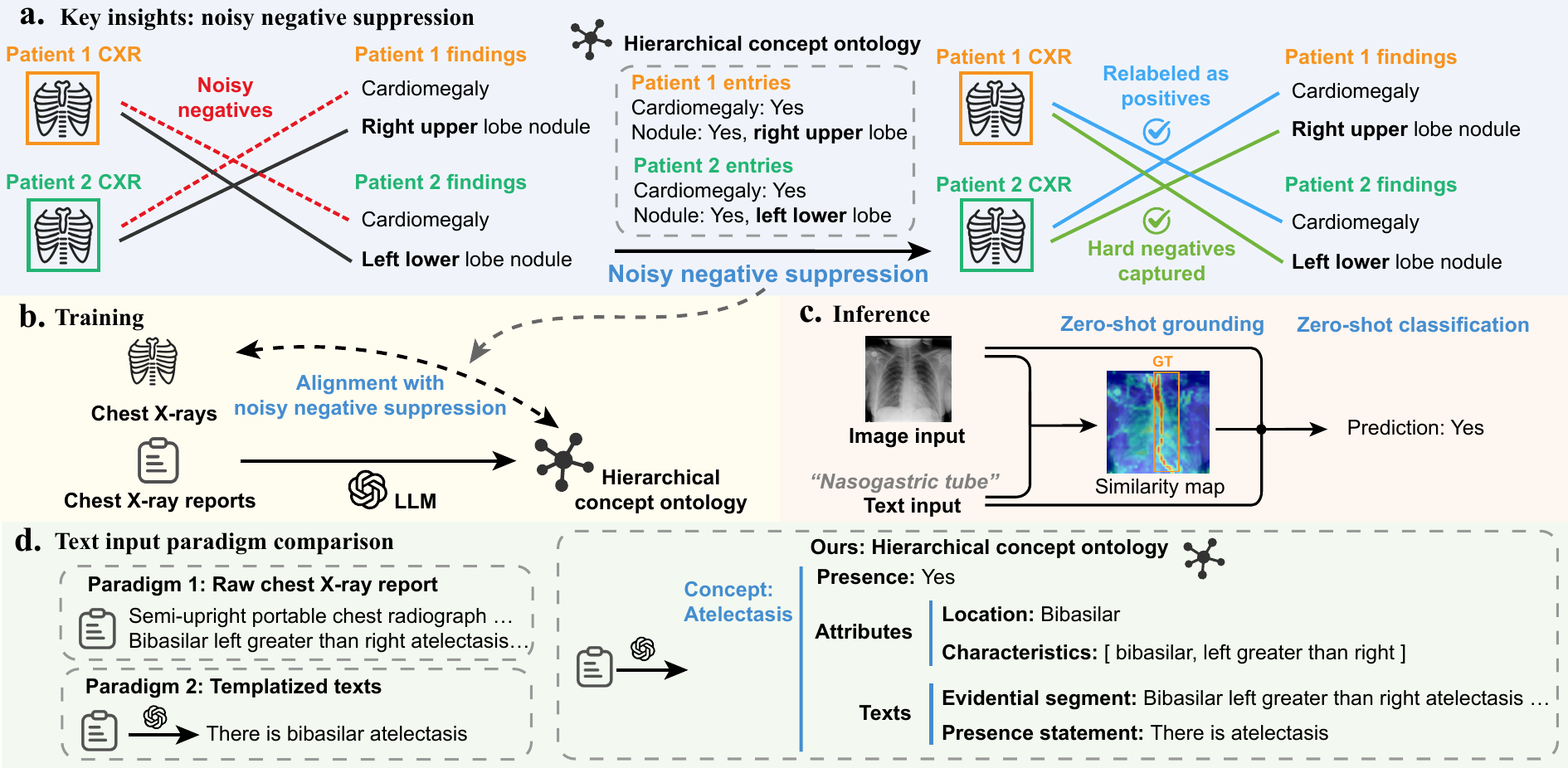}
    \caption{
    \textbf{a.} Noisy negatives and the proposed suppression strategy.
    \textbf{b.} Training and \textbf{c.} inference flows.
    \textbf{d.} Comparison of two mainstream paradigms of text inputs with ours.
    }
    \label{fig:intro}
\end{figure}

Vision-language alignment using chest X-rays (CXR) and radiology reports significantly advances zero-shot classification and grounding of radiological findings without the need for costly manual annotations (e.g., class labels or bounding boxes)~\cite{lai2024carzero,lian2025efficient,parkradzero,radford2021learning,zhang2022contrastive}.
Despite these advancements, standard image-text contrastive learning in radiology suffers from a critical, often overlooked limitation: \textbf{\textit{noisy negatives}}.
The conventional CLIP-style training objective naively treats image-text pairs from different patients as negative samples~\cite{radford2021learning}.
However, in radiology datasets, distinct patients frequently exhibit identical findings~\cite{wang2022medclip}.
As shown in Fig.~\ref{fig:intro}a, Patient 1 and Patient 2 both present ``cardiomegaly''.
Forcing these semantically equivalent pairs apart as negatives introduces contradictory supervision signals, causing semantic ambiguity that hinders the learning of discriminative features.
Alongside the noisy negative issue, there exists a dilemma in utilizing text supervision.
While previous work has observed that using template text (e.g., ``There is \texttt{[finding]}'') instead of raw reports improves prompt consistency between training and inference stages~\cite{lai2024carzero}, this reduction in linguistic diversity impairs the understanding of complex, naturalistic clinical texts.

To bridge these gaps, we propose \ours, a concept-guided framework to suppress noisy negative pairs while simultaneously preserving clinical report diversity and prompt consistency.
Fig.~\ref{fig:intro}d compares the proposed \ours with two mainstream text utilization paradigms.
Paradigm 1 utilizes raw report texts~\cite{bannur2023learning,zhou2023advancing}; while this preserves natural diversity, it suffers from poor prompt consistency between training (long, dense reports) and inference (short, atomic queries).
Paradigm 2 employs fixed-style templates~\cite{lai2024carzero,parkradzero}, which ensures prompt consistency but sacrifices the linguistic diversity required to handle complex cases.
Instead of relying on raw text or rigid templates, we construct a \textbf{\textit{hierarchical concept ontology}} derived from radiology reports.
The ontology decomposes reports into structured knowledge, explicitly modeling presence, attributes (location and characteristics), and texts (evidential segments and presence statements).
The extracted evidential segments preserve linguistic diversity, while the standardized presence statements ensure prompt consistency.

Building on this concept ontology, we introduce a \textbf{\textit{cross-patient pair relabeling}} strategy incorporating three steps: (1) \textit{Fine-Grained Breakdown} to categorize pairs based on finding presence; (2) \textit{Noisy Negative Filtering} to resolve semantic conflicts by removing false negatives; and (3) \textit{Hard Negative Mining} to identify subtle attribute discrepancies using a lightweight language model.
Finally, to achieve vision-language alignment using these relabeled pairs, we propose a \textbf{\textit{Concept-Aware NCE Loss}}, which aligns visual features with all semantically consistent reports in the batch while masking out noisy negatives.

\noindent{\textbf{Main contributions.}}~\textbf{1.} We address the critical challenge of \textit{noisy negatives} in radiology vision-language alignment, proposing a new learning paradigm that shifts from patient-based to concept-based negative pair identification.
\textbf{2.} The proposed \textit{hierarchical concept ontology} simultaneously preserves clinical text diversity and ensures prompt consistency.
\textbf{3.} We propose a \textit{Concept-Aware NCE loss} tailored for the concept-based framework, suppressing noisy negatives during vision-language alignment.
\textbf{4.} Extensive experiments across multi-granularity (word, phrase, and sentence levels) grounding tasks and five zero-shot classification datasets validate that \ours outperforms previous state-of-the-art methods.
\section{Methodology}
\begin{figure}[!t]
    \centering
    \includegraphics[width=\linewidth]{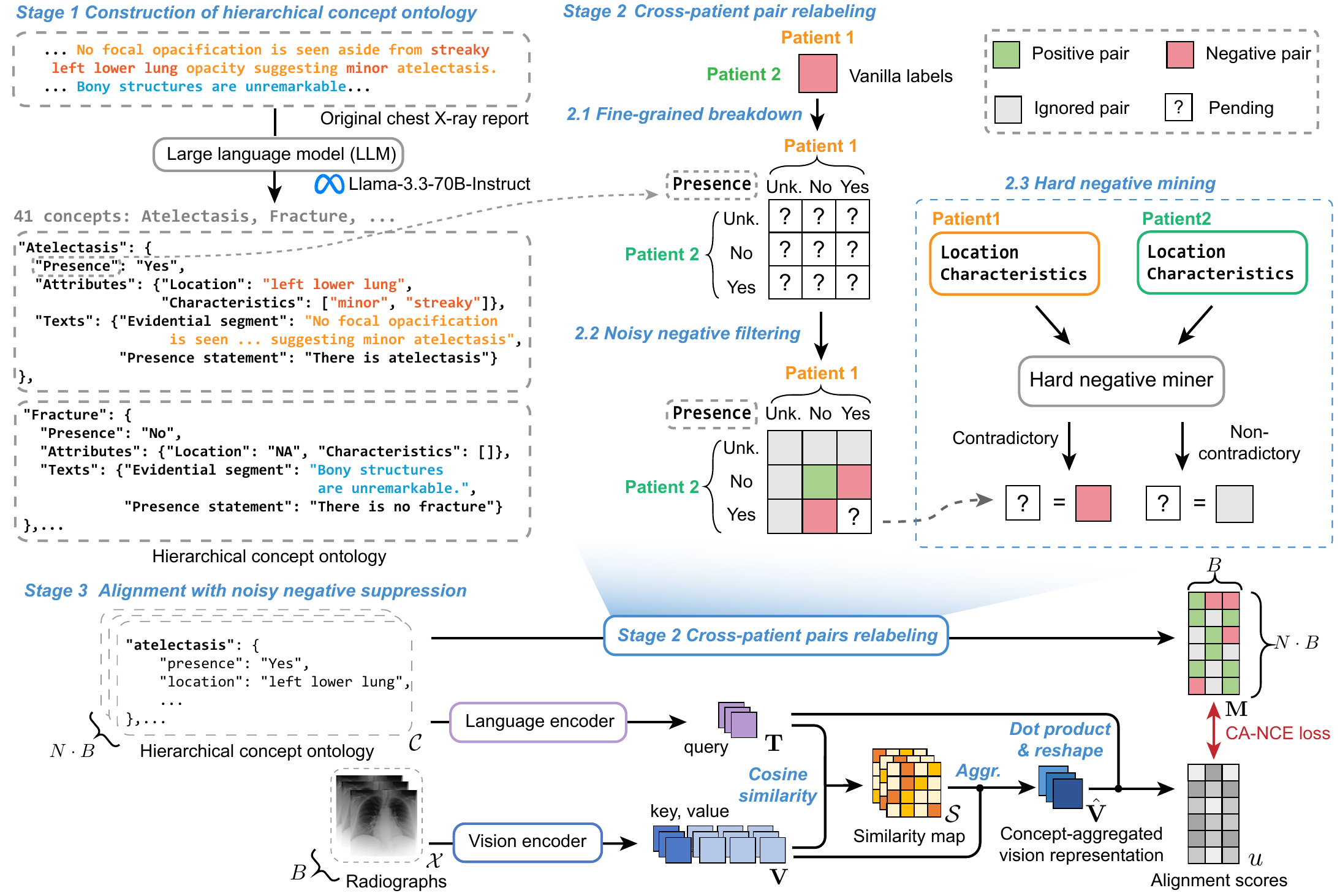}
    \caption{The proposed \ours consists of three stages.
    (1) We construct a hierarchical concept ontology via LLM.
    (2) Cross-patient pairs are relabeled into a relation matrix based on the ontology (Unk. = Unknown).
    (3) We perform vision-language alignment with noisy negative suppression using the proposed Concept-Aware NCE loss.
    }
    \label{fig:framework}
\end{figure}
\noindent{\textbf{Overview.}}~As illustrated in Fig.~\ref{fig:framework}, our framework comprises three primary stages: 1. \textbf{\textit{construction of hierarchical concept ontology}} for 41 key clinical concepts (Sec.~\ref{sec:ontology}); 2. \textit{\textbf{cross-patient pair relabeling}}, which generates a relation matrix for image-text pairs based on the hierarchical concept ontology (Sec.~\ref{sec:relabeling}); and 3. \textit{\textbf{vision-language alignment with noisy negative suppression}}, using the proposed Concept-Aware NCE (CA-NCE) loss (Sec.~\ref{sec:alignment}).
\subsection{Construction of Hierarchical Concept Ontology}
\label{sec:ontology}
To facilitate the proposed noisy negative suppression, we construct a hierarchical concept ontology derived from radiology reports in MIMIC-CXR~\cite{johnson2019mimic}.
Utilizing locally deployed Llama-3.3-70B-Instruct~\cite{dubey2024llama}, we process the reports to extract information for 41 key clinical concepts.
As illustrated in Fig.~\ref{fig:framework} (Stage 1), the ontology decomposes each finding into \textbf{\textit{three layers}}:
(1) 41 key clinical concepts of chest X-ray findings, such as ``atelectasis''.
(2) Components including \texttt{Presence}, \texttt{Attributes}, and \texttt{Texts}, where \texttt{Presence} guides the fine-grained breakdown in Stage 2.1 and noisy negative filtering in Stage 2.2;  \texttt{Attributes} are used for hard negative mining in Stage 2.3; while \texttt{Texts} serve as text inputs in Stage 3.
(3) Granular details, where \texttt{Attributes} include \texttt{Location} and \texttt{Characteristics}, while \texttt{Texts} comprise \texttt{Evidential segment} and \texttt{Presence statement} (each sampled with a 50\% probability during training).
\subsection{Cross-Patient Pair Relabeling}
\label{sec:relabeling}
Following standard practice, we regard pairs consisting of images and text from the same patient as positive samples.
Regarding image-text pairs from different patients, standard CLIP-style training assumes them to be negative samples.
However, this assumption often fails in radiology, where descriptions like ``No pneumothorax'' are frequently shared by different patients, resulting in \textbf{\textit{noisy negatives}}.
To mitigate this, we propose a relabeling strategy to construct a \textit{\textbf{relation matrix}} $\mathbf{M} \in \{1, 0, \varnothing\}^{(N \cdot B) \times B}$, where $N$ and $B$ refer to the number of sampled texts per image (default 8) and the batch size (default 192), respectively.
Entries in $\mathbf{M}$ represent positives ($1$), negatives ($0$), and ambiguous pairs ($\varnothing$).
The construction of $\mathbf{M}$ proceeds in three steps:

\noindent{\textbf{1. Fine-grained breakdown.}}~
We first construct a presence-based $3 \times 3$ grid by pairing three possible statuses: ``Yes'' (present), ``No'' (absent), or ``Unk.'' (Unknown) (Fig.~\ref{fig:framework}, Stage 2.1), which provides the foundation for subsequent logic.

\noindent{\textbf{2. Noisy negative filtering.}}~
Based on the breakdown, we categorize the pairs into four initial types and populate the relation matrix $\mathbf{M}$ (Fig.~\ref{fig:framework}, Stage 2.2):

\noindent$\sbullet$Positive pairs ($\mathbf{M}_{i,j} = 1$): ``No-No'' pairs, a consistent absence of the concept.

\noindent$\sbullet$Negative pairs ($\mathbf{M}_{i,j} = 0$): ``Yes-No'' pairs, where the presence is contradictory.

\noindent$\sbullet$Ignored pairs ($\mathbf{M}_{i,j} = \varnothing$): Pairs containing an ``Unknown'' status are considered semantically ambiguous and are therefore ignored when computing loss.

\noindent$\sbullet$Pending pairs: ``Yes-Yes'' pairs, where both patients share the finding but may differ in characteristics (e.g., small vs. large) and location (e.g., left vs. right). These are deferred to the next stage to avoid introducing noise.

\noindent{\textbf{3. Hard negative mining.}}~
To facilitate fine-grained discrimination, we tackle the pending ambiguous pairs using \texttt{Location} and \texttt{Characteristics} (Fig.~\ref{fig:framework}, Stage 2.3).
We compare the attributes of Patient 1 against Patient 2 using nli-deberta-v3-small~\cite{reimers-2019-sentence-bert}.
If the attributes are explicitly contradictory (e.g., Left vs. Right), the pair is relabeled as a \textbf{hard negative} ($\mathbf{M}_{i,j} = 0$).
Otherwise, if attributes are similar or insufficient to distinguish, we treat them as potential noisy negatives and mask them out ($\mathbf{M}_{i,j} = \varnothing$) to avoid semantic ambiguity.
\subsection{Vision-Language Alignment with Noisy Negative Suppression}
\label{sec:alignment}
Building upon the negative pair relabeling in Sec.~\ref{sec:relabeling}, we introduce the \textbf{Concept-Aware NCE (CA-NCE)} loss.
Guided by the relation matrix $\mathbf{M}$, this loss function aligns concept-aggregated visual representations with corresponding texts.

\noindent{\textbf{Counterfactual prompting.}}~To enhance the diversity of negative image-text pairs, we introduce counterfactual texts for each target concept by sampling descriptions from patients with the opposite \texttt{Presence} status.
In these cases, fine-grained attributes (i.e., \texttt{location} and \texttt{characteristics}) are set to null to prevent the introduction of spurious correlations during hard negative mining.

\noindent{\textbf{Vision and text encoding.}}~Let $\mathcal{X} = \{x_j\}_{j=1}^B$ be a batch of chest X-rays and $\mathcal{C} = \{c_i\}_{i=1}^{N \cdot B}$ be the concept-centric input texts.
We employ a dual-encoder architecture with a frozen Rad-DINO-Maira-2~\cite{bannur2024maira,perez2025exploring} followed by two trainable ViT layers~\cite{dosovitskiy2020image} as the vision encoder (input size $518\times 518$), and BiomedVLP-CXR-BERT~\cite{boecking2022making} as the text encoder (length of 128 tokens).
Inputs are processed to obtain the visual and textual representations: $\mathbf{V} = f_{vis}(\mathcal{X})$ and $\mathbf{T} = f_{txt}(\mathcal{C})$.
Here, $\mathbf{V} \in \mathbb{R}^{B \times L \times D}$ denotes the batch of image patch embeddings, where $\mathbf{v}_{j,l} \in \mathbb{R}^D$ represents the raw feature vector of the $l$-th patch in the $j$-th image ($j \in \{1, \dots, B\}, l \in \{1, \dots, L\}$).
Correspondingly, $\mathbf{T} \in \mathbb{R}^{(N\cdot B) \times D}$ denotes the text embeddings, where $\mathbf{t}_i \in \mathbb{R}^D$ corresponds to the $i$-th text ($i \in \{1, \dots, (N\cdot B)\}$).

\noindent{\textbf{Similarity map and concept-aggregated vision representation.}}~
Let $\bar{\mathbf{t}}_i$ and $\bar{\mathbf{v}}_{j,l}$ denote the $\ell_2$-normalized text and image patch embeddings, respectively.
To achieve concept-aware fine-grained alignment, we first compute the similarity map $\mathcal{S}_{i,j} \in \mathbb{R}^L$, which consists of patch-wise similarity scores $s_{i,j,l}$, defined as the scaled dot-product: $s_{i,j,l} = (\bar{\mathbf{t}}_i^\top \bar{\mathbf{v}}_{j,l}) / \tau_{attn}$.
Subsequently, we normalize these scores via softmax to generate attention weights $w_{i,j,l}$, which are used to aggregate the visual features into a concept-specific representation $\hat{\mathbf{v}}_{i,j}$ ($\hat{\mathbf{V}}$ in Fig.~\ref{fig:framework} refers to representations $\{\hat{\mathbf{v}}_{i,j}\}$ in a batch):
\begin{equation}
    w_{i,j,l} = \frac{\exp(s_{i,j,l})}{\sum_{k=1}^{L} \exp(s_{i,j,k})}, \quad \mathbf{z}_{i,j} = \sum_{l=1}^{L} w_{i,j,l} \bar{\mathbf{v}}_{j,l}, \quad \hat{\mathbf{v}}_{i,j} = \frac{\mathbf{z}_{i,j}}{\|\mathbf{z}_{i,j}\|_2}.
    \label{eq:aggregation}
\end{equation}
The final alignment score $u_{i,j}$ is computed as $u_{i,j} = \bar{\mathbf{t}}_i^\top \hat{\mathbf{v}}_{i,j}$.

Notably, for zero-shot grounding in inference, $\mathcal{S}_{i,j}$ is reshaped to the spatial grid size $H \times W$ (where $H \cdot W=L$) and interpolated to the original image resolution to serve as the localization heatmap.

\noindent{\textbf{Concept-aware NCE loss.}}~
To suppress noisy negatives identified in the relabeling stage and accommodate arbitrary numbers of positive pairs, we propose Concept-Aware NCE loss based on InfoNCE loss~\cite{oord2018representation}, which incorporates a scaling temperature $\tau_{loss}$.
We first derive binary masks from the relation matrix $\mathbf{M}$ constructed in Sec.~\ref{sec:relabeling}.
The \textbf{positive mask} $\mathbf{M}_{pos}$ indicates positive pairs, and the \textbf{valid mask} $\mathbf{M}_{valid}$ indicates pairs participating in the loss:
\begin{equation}
    \mathbf{M}_{pos}(i,j) = \mathbb{I}(\mathbf{M}_{i,j} = 1), \quad
    \mathbf{M}_{valid}(i,j) = \mathbb{I}(\mathbf{M}_{i,j} \neq \varnothing).
\end{equation}
The Text-to-Image ($\mathcal{L}_{T2I}$) and Image-to-Text ($\mathcal{L}_{I2T}$) losses are defined as:
\begin{subequations}
\begin{align}
    \mathcal{L}_{T2I}(i) &= - \log \left( \frac{\sum_{j=1}^{B} \mathbf{M}_{pos}(i,j) \cdot \exp(u_{i,j} / \tau_{loss})}{\sum_{k=1}^{B} \mathbf{M}_{valid}(i,k) \cdot \exp(u_{i,k} / \tau_{loss}) + \epsilon} \right),\\
    \mathcal{L}_{I2T}(j) &= - \log \left( \frac{\sum_{k=1}^{N\cdot B} \mathbf{M}_{pos}(k,j) \cdot \exp(u_{k,j} / \tau_{loss})}{\sum_{m=1}^{N\cdot B} \mathbf{M}_{valid}(m,j) \cdot \exp(u_{m,j} / \tau_{loss}) + \epsilon} \right).
\end{align}
\end{subequations}
The total objective is the average over entries containing positive targets:
\begin{equation}
    \mathcal{L} = \frac{\sum_{i} \mathbb{I}(p_i > 0) \mathcal{L}_{T2I}(i)}{\sum_{i} \mathbb{I}(p_i > 0)} + \frac{\sum_{j} \mathbb{I}(q_j > 0) \mathcal{L}_{I2T}(j)}{\sum_{j} \mathbb{I}(q_j > 0)}
    \label{eq:total_loss}.
\end{equation}
where $p_i = \sum_{j} \mathbf{M}_{pos}(i,j)$ and $q_j = \sum_{k} \mathbf{M}_{pos}(k,j)$.
\section{Experiments}
\subsection{Dataset and Evaluation Metrics}
\noindent{\textbf{Training dataset.}}~We employ MIMIC-CXR~\cite{johnson2019mimic} for vision-language alignment training.
This dataset comprises 227,835 radiographic studies from 65,379 patients.
We follow the official split for training and validation.
Each study consists of a radiology report and one or more chest X-ray images, totaling 377,100 images.
These reports serve as the source for constructing the hierarchical concept ontology, which is described in Sec.~\ref{sec:ontology}.

\noindent{\textbf{Evaluation datasets.}}~We conduct a comprehensive evaluation using three zero-shot grounding datasets and five zero-shot classification datasets.
For \textit{\textbf{zero-shot grounding}}, we systematically evaluate the performance of \textbf{\textit{word-level, phrase-level, and sentence-level}} grounding tasks on ChestXDet10~\cite{liu2020chestx}, MS-CXR~\cite{boecking2022making}, and PadChest-GR~\cite{de2025padchest}, respectively.
To measure grounding performance, we utilize the Pointing Game score~\cite{zhang2018top}, which calculates the hit rate where the maximum value falls within the ground-truth bounding box.
For \textbf{zero-shot classification}, we employ ChestXDet10~\cite{liu2020chestx}, Open-I~\cite{demner2015preparing}, ChestXray14~\cite{wang2017chestx}, CheXpert~\cite{irvin2019chexpert}, and PadChest-GR~\cite{de2025padchest}.
The Area Under the Receiver Operating Characteristic curve (AUROC) is adopted as the classification metric.
Specifically, the test set of \textbf{ChestXDet10} includes 542 images with manually annotated bounding boxes for 10 diseases.
\textbf{MS-CXR} contains 1,153 phrase-bounding box pairs.
For a fair comparison, evaluation is conducted on the test set released by~\cite{chen2023medical}, comprising 167 images.
\textbf{PadChest-GR} contains 915 images in the test set, highlighting box-level labels with sentence-level descriptions of 24 findings.
\textbf{Open-I} comprises 7,470 chest X-ray images with annotations for 18 diseases.
The official test set of \textbf{ChestXray14} contains 22,433 images annotated with 14 disease labels.
\textbf{CheXpert} includes 500 patients in the test set; our evaluation focuses on the 5 officially recommended diseases.
\begin{table}[!t]
\caption{
Comparison with SOTAs.
Pointing Game scores (\%) for zero-shot grounding and AUROCs (\%) for zero-shot classification are reported as mean $\pm$ std over 3 runs.
``-'': not applicable.
The best results are in \textbf{bold} and the second-best ones are \uline{underlined}.
}
\renewcommand\arraystretch{1.1}
\resizebox{\linewidth}{!}{
\begin{tabular}{c|c|ccc|ccccc}
\toprule
                         &                         & \multicolumn{3}{c|}{Zero-shot grounding}                                                              & \multicolumn{5}{c}{Zero-shot classification}                                            \\ \cline{3-10} 
\multirow{-2}{*}{Method} & \multirow{-2}{*}{Venue} & \multicolumn{1}{c|}{ChestXDet10}   & \multicolumn{1}{c|}{MS-CXR}                      & PadChest-GR   & \multicolumn{1}{c|}{ChestXDet10}   & \multicolumn{1}{c|}{Open-I}        & \multicolumn{1}{c|}{ChestXray14}   & \multicolumn{1}{c|}{CheXpert}      & PadChest-GR   \\ \hline
\rowcolor[HTML]{ECF1F9}
\ours                   & Ours                    & \multicolumn{1}{c|}{\textbf{65.6}{\scriptsize $\pm$0.2}} & \multicolumn{1}{c|}{\textbf{87.2}{\scriptsize $\pm$0.3}}               & \textbf{86.6}{\scriptsize $\pm$0.2} & \multicolumn{1}{c|}{\textbf{82.5}{\scriptsize $\pm$0.2}} & \multicolumn{1}{c|}{\textbf{87.1}{\scriptsize $\pm$0.1}} & \multicolumn{1}{c|}{\textbf{81.9}{\scriptsize $\pm$0.1}} & \multicolumn{1}{c|}{{\ul 92.0}{\scriptsize $\pm$0.2}}    & \textbf{88.1}{\scriptsize $\pm$0.2} \\ \hline
RadZero~\cite{parkradzero}                  & NeurIPS'25              & \multicolumn{1}{c|}{{\ul 62.2}}    & \multicolumn{1}{c|}{{\ul 84.4}}                  & {\ul 85.3}    & \multicolumn{1}{c|}{78.7}          & \multicolumn{1}{c|}{{\ul 84.7}}    & \multicolumn{1}{c|}{80.4}          & \multicolumn{1}{c|}{90.0}          & 84.7          \\
CARZero~\cite{lai2024carzero}                  & CVPR'24                 & \multicolumn{1}{c|}{54.3}          & \multicolumn{1}{c|}{79.4}                        & 77.1          & \multicolumn{1}{c|}{{\ul 79.6}}    & \multicolumn{1}{c|}{83.8}          & \multicolumn{1}{c|}{{\ul 81.1}}    & \multicolumn{1}{c|}{\textbf{92.3}} & {\ul 86.1}    \\
BiomedCLIP~\cite{zhang2024biomedclip}               & NEJM AI'24              & \multicolumn{1}{c|}{-}             & \multicolumn{1}{c|}{-}                           & -             & \multicolumn{1}{c|}{63.0}          & \multicolumn{1}{c|}{57.7}          & \multicolumn{1}{c|}{63.9}          & \multicolumn{1}{c|}{67.7}          & 64.7          \\
KAD~\cite{zhang2023knowledge}                      & Nat. Commun.'23         & \multicolumn{1}{c|}{39.1}          & \multicolumn{1}{c|}{19.2}                        & 29.2         & \multicolumn{1}{c|}{73.5}          & \multicolumn{1}{c|}{80.7}          & \multicolumn{1}{c|}{78.9}          & \multicolumn{1}{c|}{90.5}          & 67.3          \\
MedKLIP~\cite{wu2023medklip}                  & ICCV'23                 & \multicolumn{1}{c|}{48.1}          & \multicolumn{1}{c|}{40.7}                        & 34.8          & \multicolumn{1}{c|}{71.3}          & \multicolumn{1}{c|}{75.9}          & \multicolumn{1}{c|}{72.6}          & \multicolumn{1}{c|}{87.9}          & 67.9          \\
BioViL-T~\cite{bannur2023learning}                 & CVPR'23                 & \multicolumn{1}{c|}{35.1}          & \multicolumn{1}{c|}{{71.9}} & 22.8          & \multicolumn{1}{c|}{70.8}          & \multicolumn{1}{c|}{70.2}          & \multicolumn{1}{c|}{72.9}          & \multicolumn{1}{c|}{78.9}          & 52.9    \\
\bottomrule
\end{tabular}
}
\label{table:main}
\end{table}
\begin{table}[t]
\caption{
Pointing Game scores (\%) per category for zero-shot grounding on ChestXDet10.
The best results are in \textbf{bold} and the second-best ones are \uline{underlined}.
}
\renewcommand\arraystretch{1}
\resizebox{\linewidth}{!}{
\begin{tabular}{C{2cm}|C{2.7cm}|C{1.2cm}|*{10}{C{1cm}}}
\toprule
Method   & Venue           & Mean          & ATE           & CALC          & CONS          & EFF           & EMPH          & FIB           & FX            & MASS          & NOD           & PTX           \\ \hline
\rowcolor[HTML]{ECF1F9}
\ours   & Ours            & \textbf{65.6} & \textbf{83.3} & \textbf{44.7} & 74.4          & {\ul 80.2}    & 76.9          & \textbf{65.9} & \textbf{40.8} & 63.3          & \textbf{63.6} & \textbf{62.9} \\ \hline
RadZero~\cite{parkradzero}  & NeurIPS'25      & 62.2          & 64.6          & 36.8          & {\ul 82.4}    & \textbf{85.7} & \textbf{87.2} & 58.5          & 25.0          & \textbf{76.7} & 50.6          & 54.3          \\
CARZero~\cite{lai2024carzero}  & CVPR'24         & 54.3          & 60.4          & 18.4          & {\ul 82.4}    & 78.2          & {\ul 84.6}    & 56.1          & 18.4          & 70.0          & 28.6          & 45.7          \\
KAD~\cite{zhang2023knowledge}      & {Nat. Commun.'23} & 39.1          & 64.6          & 13.2          & 69.9          & 61.8          & 64.4          & 24.4          & 19.9          & 26.7          & 31.6          & 14.3          \\
MedKLIP~\cite{wu2023medklip}  & ICCV'23         & 48.1          & 62.5          & 13.2          & \textbf{83.7} & 67.5          & 73.4          & 30.5          & 22.4          & {\ul 73.3}    & 31.2          & 22.9          \\
BioViL-T~\cite{bannur2023learning} & CVPR'23         & 35.1          & 43.8          & 0.0           & 63.0          & 50.4          & {\ul 84.6}    & 39.0          & 2.6           & 50.0          & 0.0           & 17.1         \\
\bottomrule
\end{tabular}
}
\label{table:grounding}
\end{table}
\subsection{Results and Analyses}
\noindent{\textbf{{Comparison with the state-of-the-art.}}}
We compare the proposed \ours against state-of-the-art methods, spanning zero-shot understanding~\cite{lai2024carzero,parkradzero} and vision-language pretraining~\cite{bannur2023learning,wu2023medklip,zhang2024biomedclip,zhang2023knowledge}.
As presented in Table~\ref{table:main}, \ours achieves superior performance across both zero-shot grounding and zero-shot classification tasks.
Specifically, for zero-shot grounding, \ours significantly outperforms the runner-up RadZero~\cite{parkradzero} across all multi-granularity datasets ($p < 0.01$), achieving absolute gains of 3.4\%, 2.8\%, and 1.3\% on ChestXDet10, MS-CXR, and PadChest-GR, respectively.
A fine-grained analysis on ChestXDet10 (Table~\ref{table:grounding}) further reveals that \ours achieves the highest Pointing Game scores in 6 out of 10 categories.
Notably, in challenging categories where prior methods struggle, such as CALC and FX, we observe substantial relative improvements of 21.5\% and 63.2\%, respectively.
In Table~\ref{table:grounding}, we adopt the following abbreviations for the findings: Atelectasis (ATE), Calcification (CALC), Consolidation (CONS), Effusion (EFF), Emphysema (EMPH), Fibrosis (FIB), Fracture (FX), Mass (MASS), Nodule (NOD), and Pneumothorax (PTX).
In zero-shot classification, \ours also exhibits higher AUROCs on four out of five datasets ($p<0.01$) while achieving performance comparable to CARZero~\cite{lai2024carzero} on CheXpert ($p=0.12$).
Partial results in Table~\ref{table:main} and~\ref{table:grounding} are cited from~\cite{lai2024carzero} and~\cite{parkradzero}.

\begin{figure}[!t]
    \centering
    \includegraphics[width=\linewidth]{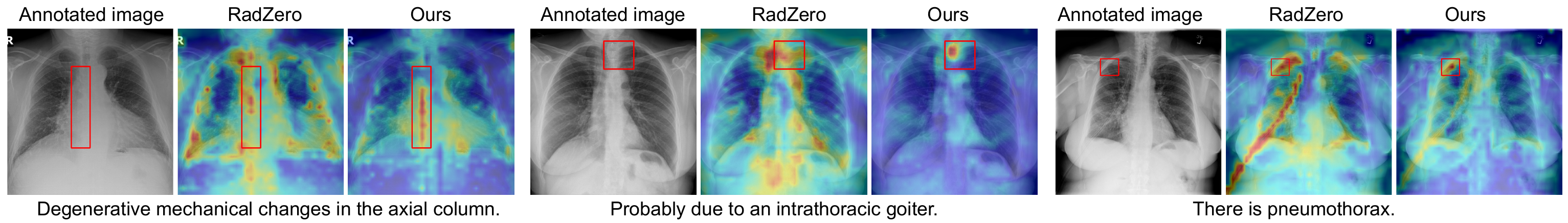}
    \caption{Qualitative comparison of similarity maps between ours and the leading competing method~\cite{parkradzero}.
    Red bounding boxes indicate ground-truth annotations.
    }
    \label{fig:visualization}
\end{figure}

\noindent{\textbf{{Qualitative analysis.}}}
To provide an intuitive comparison of grounding capabilities, we visualize the similarity maps of the proposed \ours alongside the leading state-of-the-art method, RadZero~\cite{parkradzero}.
As illustrated in Fig.~\ref{fig:visualization}, \ours generates a more concentrated focus that aligns precisely with the ground-truth bounding boxes.
These qualitative results validate that the proposed noisy negative suppression strategy effectively mitigates semantic ambiguity, enabling the precise localization of complex findings.

\begin{table}[!t]
\caption{Ablation study of main components in \ours.
Pointing Game scores and AUROCs (\%) are reported for zero-shot grounding and classification, respectively.
}
\renewcommand\arraystretch{1.1}
\resizebox{\linewidth}{!}{
\begin{tabular}{c|c|c|c|ccc|ccc} 
\toprule

\multirow{2}{*}{Setting} & \multirow{2}{*}{\begin{tabular}[c]{@{}c@{}}Noisy negative\\ filtering\end{tabular}} & \multirow{2}{*}{\begin{tabular}[c]{@{}c@{}}Hard negative\\ mining\end{tabular}} & \multirow{2}{*}{\begin{tabular}[c]{@{}c@{}}Counterfactual\\ prompting\end{tabular}} & \multicolumn{3}{c|}{Zero-shot grounding} & \multicolumn{3}{c}{Zero-shot classification} \\ \cline{5-10} 

 &  &  &  & \multicolumn{1}{c|}{ChestXDet10} & \multicolumn{1}{c|}{MS-CXR} & PadChest-GR & \multicolumn{1}{c|}{ChestXDet10} & \multicolumn{1}{c|}{Open-I} & ChestXray14 \\ \hline

(a) & \checkmark &  &  & \multicolumn{1}{c|}{61.1 (\textcolor{red}{$-$4.5})} & \multicolumn{1}{c|}{82.0 (\textcolor{red}{$-$5.2})} & 88.2 (\textcolor{blue}{$+$1.6}) & \multicolumn{1}{c|}{81.2 (\textcolor{red}{$-$1.3})} & \multicolumn{1}{c|}{84.0 (\textcolor{red}{$-$3.1})} & 81.6 (\textcolor{red}{$-$0.3}) \\

(b) &  &  & \checkmark & \multicolumn{1}{c|}{58.5 (\textcolor{red}{$-$7.1})} & \multicolumn{1}{c|}{85.6 (\textcolor{red}{$-$1.6})} & 86.2 (\textcolor{red}{$-$0.4}) & \multicolumn{1}{c|}{81.7 (\textcolor{red}{$-$0.8})} & \multicolumn{1}{c|}{86.9 (\textcolor{red}{$-$0.2})} & 82.1 (\textcolor{blue}{$+$0.2}) \\

(c) & \checkmark & \checkmark &  & \multicolumn{1}{c|}{62.6 (\textcolor{red}{$-$3.0})} & \multicolumn{1}{c|}{87.4 (\textcolor{blue}{$+$0.2})} & 87.9 (\textcolor{blue}{$+$1.3}) & \multicolumn{1}{c|}{81.4 (\textcolor{red}{$-$1.1})} & \multicolumn{1}{c|}{87.0 (\textcolor{red}{$-$0.1})} & 81.9 (\textcolor{red}{$-$0.0}) \\

(d) & \checkmark &  & \checkmark & \multicolumn{1}{c|}{61.2 (\textcolor{red}{$-$4.4})} & \multicolumn{1}{c|}{82.6 (\textcolor{red}{$-$4.6})} & 87.0 (\textcolor{blue}{$+$0.4}) & \multicolumn{1}{c|}{82.0 (\textcolor{red}{$-$0.5})} & \multicolumn{1}{c|}{84.7 (\textcolor{red}{$-$2.4})} & 81.2 (\textcolor{red}{$-$0.7}) \\ \hline

\rowcolor[HTML]{ECF1F9}
Ours & \checkmark & \checkmark & \checkmark & \multicolumn{1}{c|}{65.6} & \multicolumn{1}{c|}{87.2} & 86.6 & \multicolumn{1}{c|}{82.5} & \multicolumn{1}{c|}{87.1} & 81.9 \\ \bottomrule
\end{tabular}
}
\label{table:ablation}
\end{table}
\noindent{\textbf{{Ablation study.}}}
We conduct an ablation study to validate the contributions of the proposed noisy negative suppression (comprising noisy negative filtering and hard negative mining) alongside the counterfactual prompting strategy.
As shown in Table~\ref{table:ablation}, removing the entire noisy negative suppression leads to the sharpest performance drop (Setting b).
The consistent trend of performance degradation across all settings indicates the necessity of these core components.
\subsection{Implementation Details}
Our code is implemented using PyTorch 2.7.0~\cite{paszke2019pytorch}.
Experiments were conducted using 4 NVIDIA RTX PRO 5000 Blackwell (48 GB) GPUs.
We adopt AdamW~\cite{loshchilov2017decoupled} as the optimizer, with a weight decay of 0.05, $\beta_1$ of 0.9, and $\beta_2$ of 0.95.
A warm-up strategy is employed, where the learning rate increases linearly to 1e-4 and then decreases using a cosine decay schedule.
The model is trained for 10 epochs, during which the checkpoint yielding the lowest validation loss is selected.
We evaluate comparative methods using their official code and weights.
\section{Conclusion}
In this paper, we propose \ours, a concept-guided framework to suppress noisy negatives in chest X-ray vision-language alignment, enhancing zero-shot classification and grounding of findings.
The proposed hierarchical concept ontology and Concept-Aware NCE loss mitigate semantic ambiguity.
Comprehensive experiments validate the advancements in zero-shot tasks.
This concept-centric paradigm offers a solution for mitigating noisy negatives in medical data, which can be generalized to broader modalities in future work.
%
%
%
\bibliographystyle{splncs04}
\bibliography{mybibliography}
\end{document}